\documentclass[12pt]{article}

\usepackage{longtable}
\usepackage{booktabs}
\usepackage{newtxtext,newtxmath}
\usepackage[most]{tcolorbox}
\usepackage{graphicx}

\usepackage[letterpaper,margin=1in]{geometry}

\renewenvironment{abstract}
	{\quotation}
	{\endquotation}

\date{}

\makeatletter
\renewcommand{\fnum@figure}{\textbf{Figure \thefigure}}
\renewcommand{\fnum@table}{\textbf{Table \thetable}}
\makeatother

\usepackage{scicite}

\usepackage{url}

\def\scititle{
	Post-training makes large language models less human-like
}
\title{\bfseries \boldmath \scititle}

\author{
	Marcel~Binz$^{1\ast}$,
    Elif~Akata$^{1}$,
    Abdullah~Almaatouq$^{2}$,
    Mohammed~Alsobay$^{2}$,\and
    Oleksii~Ariasov$^{3}$,
    Franziska~Br\"andle$^{4}$,
    David~Broska$^{5}$,
    Jason~W.~Burton$^{6,7}$,\and
    Nuno~Busch$^{8}$,
    Frederick~Callaway$^{9}$,
    Vanessa~Cheung$^{10}$,
    Brian~Christian$^{4}$,\and
    Julian~Coda-Forno$^{1,8}$,
    Can~Demircan$^{1}$,
    Vittoria~Dentella$^{11}$,
    Maria~K.~Eckstein$^{12}$,\and
    No\'emi~\'Eltet\H{o}$^{13}$,
    Michael~Franke$^{3}$,
    Thomas~L.~Griffiths$^{14}$,
    Fritz~G\"unther$^{15}$,\and
    Susanne~Haridi$^{16}$,
    Sebastian~Hellmann$^{8}$,
    Stefan~Herytash$^{16}$,
    Linus~Hof$^{8}$,\and
    Eleanor~Holton$^{4}$,
    Isabelle~Hoxha$^{17,18}$,
    Zak~Hussain$^{19}$,
    Akshay~Jagadish$^{1}$,\and
    Elif~Kara$^{16}$,
    Valentin~Kriegmair$^{7}$,
    Evelina~Leivada$^{20,21}$,
    Li~Ji-An$^{22}$,
    Tobias~Ludwig$^{3}$,\and
    Maximilian~Maier$^{10, 25}$,
    Marcelo~G.~Mattar$^{9}$,
    Marvin~Mathony$^{1}$,\and
    Alireza~Modirshanechi$^{1,13}$,
    Robin~Na$^{2}$,
    Mariia~Nadverniuk$^{3}$,\and
    Antonios~Nasioulas$^{17,23}$,
    Surabhi~S.~Nath$^{13}$,
    Helen~Niemeyer$^{24}$,\and
    Kate~Nussenbaum$^{37}$,
    Sebastian~Olschewski$^{19,25}$,
    Thorsten~Pachur$^{8}$,\and
    Stefano~Palminteri$^{17}$,
    Aliona~Petrenco$^{15}$,
    Camille~V.~Phaneuf-Hadd$^{26}$,\and
    Angelo~Pirrone$^{27}$,
    Manuel~Rausch$^{28,29,30}$,
    Laura~Raveling$^{15}$,
    Shashank~Reddy$^{4,31}$,\and
    Milena~Rmus$^{1}$,
    Evan~M.~Russek$^{14}$,
    Tankred~Saanum$^{13,1}$,
    Kai~Sandbrink$^{4}$,\and
    Louis~Schiekiera$^{15,24}$,
    Johannes~A.~Schubert$^{1,4}$,
    Luca~M.~Schulze~Buschoff$^{1}$,\and
    Nishad~Singhi$^{32}$,
    Leah~H.~Somerville$^{26}$,
    Mikhail~S.~Spektor$^{33,25}$,
    Xin~Sui$^{13,3}$,\and
    Christopher~Summerfield$^{4}$,
    Mirko~Thalmann$^{1}$,
    Anna~I.~Thoma$^{7}$,\and
    Taisiia~Tikhomirova$^{7}$,
    Vuong~Truong$^{34}$,
    Polina~Tsvilodub$^{3}$,\and
    Konstantinos~Voudouris$^{1}$,
    Kristin~Witte$^{1}$,
    Shuchen~Wu$^{36}$,\and
    Dirk~U.~Wulff$^{7,19}$,
    Hua-Dong~Xiong$^{35}$,
    Songlin~Xu$^{22}$,
    Lance~Ying$^{2}$,
    Xinyu~Zhang$^{22}$,\and
    Jian-Qiao~Zhu$^{14, 38}$,
	Eric~Schulz$^{1}$\and
	\small$^{1}$Helmholtz Munich, 
	\small$^{2}$Massachusetts Institute of Technology,
	\small$^{3}$University of T\"ubingen,\\
	\small$^{4}$University of Oxford,
	\small$^{5}$Stanford University,
	\small$^{6}$University of Copenhagen,\\
	\small$^{7}$Max Planck Institute for Human Development,
	\small$^{8}$Technical University of Munich, 
	\small$^{9}$New York University,\\
	\small$^{10}$University College London,
	\small$^{11}$University of Pavia,
	\small$^{12}$Google DeepMind,\\
	\small$^{13}$Max Planck Institute for Biological Cybernetics,
	\small$^{14}$Princeton University,
	\small$^{15}$Humboldt-Universit\"at zu Berlin,\\
	\small$^{16}$LMU Munich,
	\small$^{17}$\'Ecole normale sup\'erieure,
	\small$^{18}$Leiden University,
	\small$^{19}$University of Basel,\\
	\small$^{20}$Autonomous University of Barcelona,
	\small$^{21}$Instituci\'o Catalana de Recerca i Estudis Avan\c{c}ats (ICREA),\\
	\small$^{22}$University of California San Diego,
	\small$^{23}$Paris School of Economics,
	\small$^{24}$Freie Universit\"at Berlin,\\
	\small$^{25}$University of Warwick,
	\small$^{26}$Harvard University,
	\small$^{27}$University of Liverpool,
	\small$^{28}$Hochschule Rhein-Waal,\\
	\small$^{29}$Katholische Universit\"at Eichst\"att-Ingolstadt,
	\small$^{30}$Alpe-Adria-Universit\"at Klagenfurt,\\
	\small$^{31}$Singapore Management University,
	\small$^{32}$TU Darmstadt,
	\small$^{33}$VinUniversity,
	\small$^{34}$Taipei Medical University,\\
	\small$^{35}$Georgia Institute of Technology,
	\small$^{36}$Allen Institute,
    \small$^{37}$Boston University,
    \small$^{38}$The University of Hong Kong,\\
    \small$^{39}$Hunter College, City University of New York.
}

\begin{document} 

\maketitle

\newpage
\begin{abstract} \bfseries \boldmath
Large language models (LLMs) are increasingly used as surrogates for human participants, but it remains unclear which models best capture human behavior and why. To address this, we introduce \textsc{Psych-201}, a novel dataset that enables us to measure behavioral alignment at scale. We find that post-training -- the stage that turns base models into useful assistants -- consistently reduces alignment with human behavior across model families, sizes, and objectives. Moreover, this misalignment widens in newer model generations even as base models continue to improve. Finally, we find that persona-induction -- a popular technique for eliciting human-like behavior by conditioning models on participant-specific information -- does not improve predictions at the level of individuals. Taken together, our results suggest that the very processes that are currently employed to turn LLMs into useful assistants also make them less accurate models of human behavior.
\end{abstract}

\noindent

Large language models (LLMs) such as ChatGPT, Claude, and Gemini have rapidly transformed the landscape of science and society, serving as powerful tools for writing, coding, and reasoning \cite{bommasani2021opportunities, binz2025should}. Most current development is geared toward turning these models into useful assistants that provide normatively correct responses. Yet one of their most far-reaching applications lies elsewhere: faithfully mimicking human behavior, including its errors, variance, and the factors that shape it \cite{park2023generative, park2024generative, binz2025foundation, cui2025large, hullman2026human}. These human-like models could be applied to simulate patient responses in mental health care, which, for instance, could train psychiatrists on challenging clinical cases in-silico. They could make it possible to anticipate how individuals and populations will respond to policy interventions even before those interventions are deployed. They could help model student learning trajectories in educational settings, thereby guiding the design of more effective and personalized curricula. 

However, the extent to which LLMs actually resemble human behavior remains disputed. While some studies report strong behavioral alignment between model outputs and human responses \cite{aher2023using, marjieh2024large, hu2024language, lampinen2024language}, others find more mixed results or even major divergences \cite{binz2023using, hagendorff2023human, chen2023emergence, gao2025take}.\footnote{Throughout this article, we use the term behavioral alignment to refer to the behavioral similarity between models and humans, which is distinct from (but related to) the broader usage of alignment in safety research, where it refers to model compliance with human preferences and values.} This raises fundamental questions: Which model properties drive alignment with human behavior? In which domains do they accurately capture human responses? Can they adapt to the characteristics of individual people?

To answer these questions, we need a testbed that covers a broad spectrum of behaviors. The present paper introduces such a resource in the form of \textsc{Psych-201}, a large-scale dataset of natural language transcripts from behavioral experiments. \textsc{Psych-201} was collected through an open research collaboration and crowdsourced effort, resulting in a dataset $3.5\times$ larger than its predecessor \textsc{Psych-101}, a more diverse participant population, and broader coverage of experimental paradigms. 

Equipped with this, we then evaluated a wide range of LLMs on their ability to predict human responses in \textsc{Psych-201}. These models come in different shapes and forms. Base models are the output of pretraining, in which the model learns to predict the next word in large text corpora. Post-trained models take a base model and further adapt it toward more specific objectives, such as instruction-tuning (teaching models to follow user requests), reasoning (training models to produce normatively correct responses alongside step-by-step reasoning traces), or vision (extending models to process images in addition to text). We find that:
\begin{enumerate}
    \item Post-training consistently reduces human-likeness. This effect holds across model families and applies to all post-training objectives, including instruction-tuning, reasoning, and vision. 
    \item Base models continue to improve across generations, i.e. newer models are generally more aligned. 
    \item Post-training misalignment -- defined as the alignment difference between a base model and its post-trained counterpart -- widens in newer models. 
    \item The largest post-training misalignment occurs in the domains of psycholinguistics and reasoning. 
    \item Persona-induction, a popular technique for eliciting more human-like behavior by conditioning models on participant-specific information, does not improve predictions at the level of individuals.
\end{enumerate}

Taken together, our findings have important implications for using LLMs as behavioral surrogates. Most prior studies rely on post-trained models, given their accessibility, user-friendliness, and ease of prompting. Yet our results suggest that this practice is suboptimal if the goal is to produce systems that are similar to humans, and they motivate new approaches to post-training that preserve the behavioral alignment found in base models. Looking beyond the presented results, \textsc{Psych-201} offers a useful resource for meta-analyses, automated scientific discovery, and the development of large-scale cognitive models.

\begin{figure}[!htbp]
    \centering
    \begin{tcolorbox}[
        enhanced,
        width=0.9\linewidth,
        colback=gray!5,
        colframe=black,
        boxrule=0.6pt,
        arc=2mm,
        left=7mm,right=2mm,top=2mm,bottom=0mm,
        overlay={
            \node[anchor=north west, font=\bfseries\small]
            at ([xshift=-6mm,yshift=6mm]frame.north west) {(a)};
        }
    ]
    \small
    
    In the following study, you will see complex words such as \textit{airport} or
    \textit{warzone} in the browser window. You will have to judge for each word
    whether it exists as an English word or not.
    
    You will see these words one after another. If the word presented to you exists,
    press the \textbf{E} key on your keyboard. If the word does not exist, press the
    \textbf{R} key.
    
    \vspace{0.75em}
    
    \begin{tabular}{ll}
    \texttt{descentmajor} & You press \texttt{<<R>>}. \\
    \texttt{bunkpunk}     & You press \texttt{<<R>>}. \\
    \texttt{seaplane}     & You press \texttt{<<R>>}. \\
    \texttt{bloodshed}    & You press \texttt{<<E>>}. \\
    \texttt{westscheme}   & You press \texttt{<<R>>}. \\
    \end{tabular}
    \end{tcolorbox}

    \vspace{0.5em}

    \begin{tikzpicture}
        \node[anchor=south west, inner sep=0] (img) at (0,0) {
            \includegraphics[width=\linewidth]{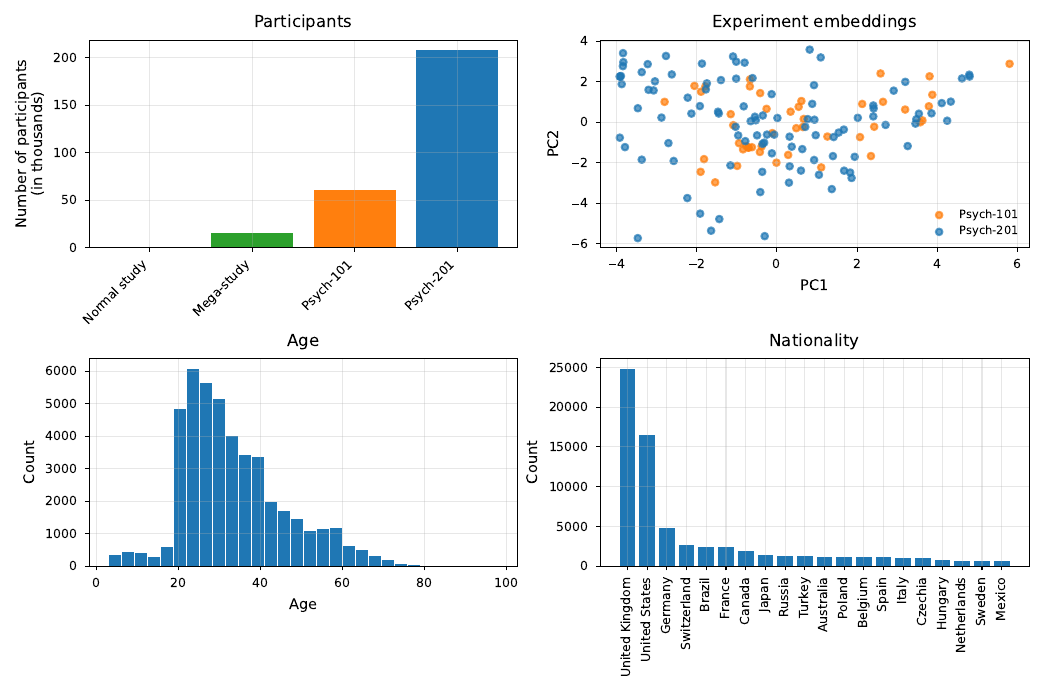}
        };

        \begin{scope}[x={(img.south east)}, y={(img.north west)}]
            \node[anchor=north west, font=\bfseries\small] at (0.01,1.0) {(b)};
            \node[anchor=north west, font=\bfseries\small] at (0.51,1.0) {(c)};
            \node[anchor=north west, font=\bfseries\small] at (0.01,0.54) {(d)};
            \node[anchor=north west, font=\bfseries\small] at (0.51,0.54) {(e)};
        \end{scope}
    \end{tikzpicture}
    \caption{\textsc{Psych-201} data example and statistics. (a) Example of a single data point, i.e. a transcript of one experimental session. (b) Number of participants across datasets, showing that \textsc{Psych-201} reaches a scale multiple times larger than typical behavioral datasets \cite{kastrati2024agreement}. (c) BERT \cite{warner2025smarter} embeddings for experiments from \textsc{Psych-101} (orange) and \textsc{Psych-201} (blue). Embeddings were projected onto two dimensions using principal component analysis. \textsc{Psych-201} covers a substantially broader spectrum of experimental paradigms than its predecessor. (d) Histogram over age in \textsc{Psych-201}. (e) Histogram over nationality in \textsc{Psych-201}.}
    \label{fig:fig1}
\end{figure}

\subsection*{\textsc{Psych-201}}

To systematically evaluate the behavioral alignment of LLMs with human responses, we introduce \textsc{Psych-201}, a novel dataset consisting of natural language transcripts from behavioral experiments. \textsc{Psych-201} contains trial-by-trial data from individual participants, with each data sequence corresponding to the transcript of an entire experimental session (including instructions, stimuli, responses, and any task-relevant context presented during the session; see Fig.~\ref{fig:fig1}a for an example).

\textsc{Psych‑201} was collected through an open research collaboration, resulting in a dataset of unprecedented scale and diversity. It includes data from $208,021$ participants, $25,906,599$ behavioral responses, and hundreds of experiments, making it $3.5\times$ larger than its predecessor \textsc{Psych‑101} and $13\times$ larger than the average mega-study \cite{kastrati2024agreement}; see Fig.~\ref{fig:fig1}b. It is furthermore more diverse than \textsc{Psych‑101}, both in terms of experimental paradigms (see Fig.~\ref{fig:fig1}c) and participant demographics. Notably, it incorporates several developmental and cross‑cultural studies, thereby increasing demographic coverage (see Fig.~\ref{fig:fig1}d-e). Each data point is annotated with detailed meta‑data, including age, nationality, questionnaire responses, and other markers.


\subsection*{Post-training makes LLMs less human-like}

We evaluated models from three major families on \textsc{Psych-201}: Qwen3, a state-of-the-art open-source model family \cite{yang2025qwen3}; Llama3.X, a model family with a broad ecosystem \cite{grattafiori2024llama}; and Olmo3.X, a fully open and reproducible model family \cite{olmo2025olmo}. For each model, we measured behavioral alignment using the negative log-likelihood (NLL) of individual human responses under predictions of the respective model (with lower values indicating closer alignment with human behavior). To quantify the effect of post-training, we report effect sizes (Cohen’s $d$) relative to the corresponding base model, averaged across all experiments unless otherwise noted. Likewise, we measure the benefit of meta-data through effect sizes between models prompted with and without participant-specific meta-data.

\begin{figure}
    \centering
    \scalebox{0.95}{\includegraphics[]{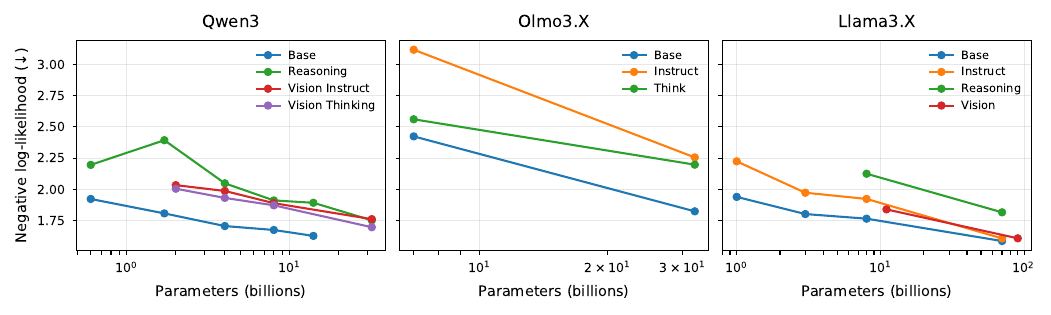}}
    \caption{Behavioral alignment for different post-training objectives. Negative log-likelihood of human responses from \textsc{Psych-201} for models from the Qwen3, Olmo3.X, and Llama3.X families (lower means more aligned). Post-trained models (shown in non-blue) exhibit consistently lower alignment with human responses across all objectives, families, and sizes than their base model counterpart (shown in blue).}
    \label{fig:fig2}
\end{figure}

We observed a consistent and robust effect across model families and sizes: post-training reduces behavioral alignment (see Fig.~\ref{fig:fig2}). This effect holds across all major post-training objectives. In the aggregate, instruction-tuned models (average $d = 0.11$), reasoning models (average $d = 0.14$), and vision models (average $d = 0.07$) are all less aligned than their corresponding base models. Furthermore, the base model outperforms its post-trained counterpart in almost every direct comparison. The most aligned model overall was Qwen2.5‑72B (average NLL = $1.557$; note that base models are not available in larger sizes for the latest Qwen generations). We present additional analysis in Fig.~\ref{fig:s1}, showing this post-training effect is robust across different prompt formats.

One potential explanation for this effect is that post-trained models simply produce more deterministic outputs, thereby failing to capture the noisiness of human behavior \cite{kirk2023understanding}. If this were the case, however, we would expect post-trained models to exhibit equal or higher accuracy since the (unchanged) mode of the output distribution would still align with human responses. Further analyses presented in Fig.~\ref{fig:s2} indicate that this is not the case, thereby ruling out increased determinism as a sole driver of the observed misalignment.

\subsection*{Behavioral alignment across model generations}

\begin{figure}[t]
    \centering
    \begin{tikzpicture}
        \node[anchor=south west, inner sep=0] (img) at (0,0) {
            \includegraphics[width=0.95\linewidth]{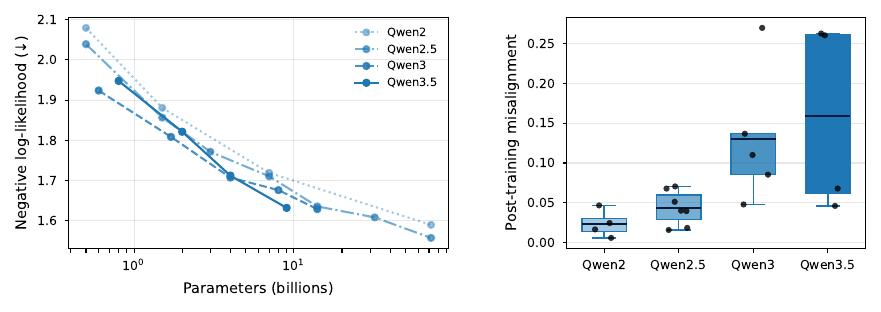}
        };

        \begin{scope}[x={(img.south east)}, y={(img.north west)}]
            \node[anchor=north west, font=\bfseries\small] at (-0.04,1.07) {(a)};
            \node[anchor=north west, font=\bfseries\small] at (0.52,1.07) {(b)};
        \end{scope}
    \end{tikzpicture}
    \caption{Behavioral alignment across Qwen generations. (a) Behavioral alignment for base models improves across generations. (b) The post-training misalignment between instruction-tuned models and their corresponding base model increases across generations.}
    \label{fig:fig3}
\end{figure}

How does behavioral alignment change with newer model generations? This is especially interesting as in some domains there is evidence that alignment with human behavior plateaus -- or even declines -- as models become more powerful \cite{linsley2023performance, oh2023does, wu2025large, collins2025evaluating}. We find that the base models, in fact, continue to improve in their behavioral alignment across model generations (see Fig.~\ref{fig:fig3}a for models from the Qwen family). However, we also find that the post-training misalignment gap widens in newer model generations (see Fig.~\ref{fig:fig3}b). For instance, the gap for instruction-tuned models is relatively modest in Qwen2 (average $d = 0.02$) and Qwen2.5 (average $d = 0.04$), but increases in Qwen3 (average $d = 0.13$) and further in Qwen3.5 (average $d = 0.16$), suggesting that ongoing developments in post-training amplify the divergence from human-like behavior.

\subsection*{Mapping behavioral alignment across experimental domains}

\begin{figure}[!htbp]
    \centering
    \scalebox{0.95}{\includegraphics[]{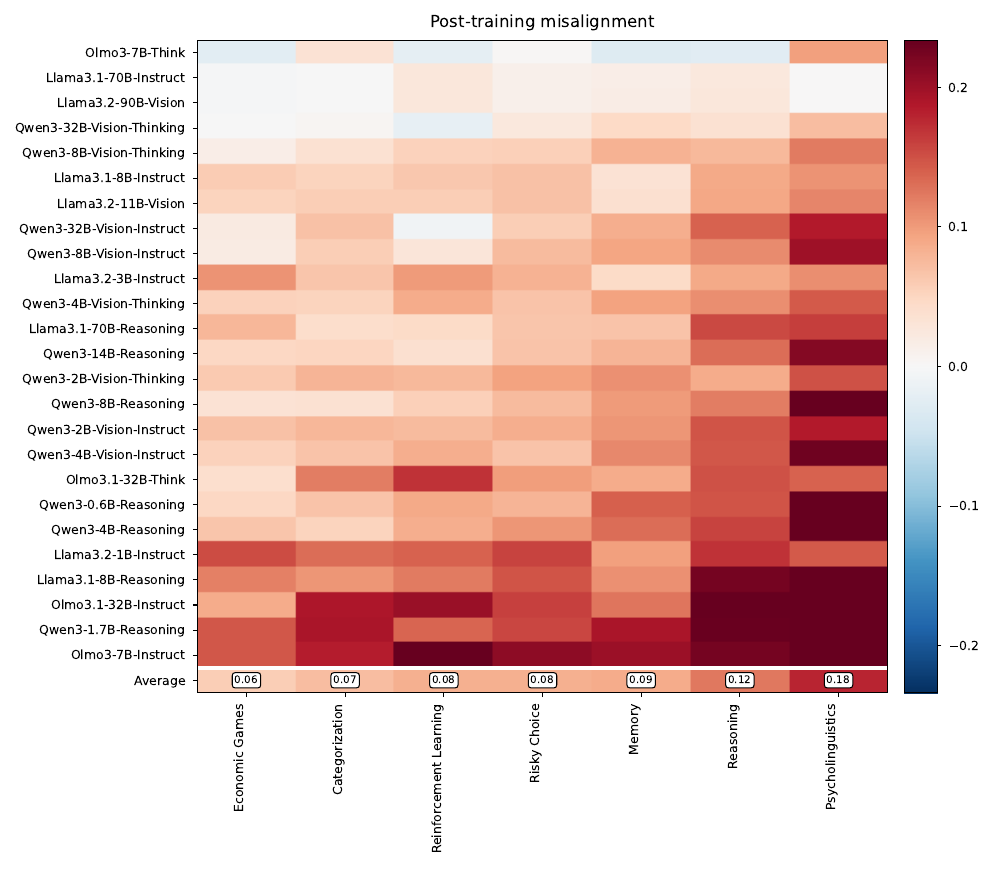}}
    \caption{Post-training misalignment by experiment domain. Each individual tile shows average effect sizes (grouped by domain) for a given model. Positive values indicate a negative effect of post-training.}
    \label{fig:fig4}
\end{figure}

Is there any structure in which types of experimental characteristics lead to more or less behavioral alignment? To investigate this, we grouped the experiments in \textsc{Psych-201} by domain and analyzed the post-training misalignment. We find that post-training misalignment is present across all domains, indicating that it is a domain-general effect (see Fig.~\ref{fig:fig4}). However, the magnitude of misalignment varies across domains and models. 

The domains with the highest post-training misalignment are reasoning and psycholinguistics. A natural explanation for this is that base models and post-trained models are optimized for different objectives: base models are ultimately models of human language, and should therefore be especially well aligned with human behavior on psycholinguistic tasks. Post-training techniques, such as reinforcement learning from human feedback, on the other hand, are designed to maximize user engagement, thereby shifting models away from their original objective. The same may hold for reasoning tasks: human decision-making is shaped by heuristics and biases \cite{gigerenzer2011heuristic, binz2022heuristics}, which might be captured by base models but are then overwritten by reasoning post-training, which optimizes for normatively correct responses. Thus, the very processes that are currently employed to turn these models into useful assistants may also make them less accurate models of human behavior.

\subsection*{No benefit of persona-induction}

Persona-induction, i.e. conditioning a model on information about a particular individual, has become a popular approach for eliciting more human-like behavior from LLMs \cite{salewski2023context, wu2026humanlm, paglieri2026persona, ma2026synthetic}. While previous studies have shown that persona-induction can produce human-aligned responses at the population level \cite{argyle2023out, park2023generative, park2024generative}, it has not been systematically evaluated whether it actually improves the prediction of responses at the level of individuals. \textsc{Psych-201}'s rich participant-level meta-data enables the first systematic test of persona-induction at scale. 

Following prior work, we adopted an interview-style prompting format \cite{lutz2025prompt} in which each experimental transcript was preceded by question-answer pairs describing the participant (e.g. “Interviewer: What is your age? Interviewee: 35.”). We included information about age, gender, nationality, education, clinical diagnosis, and questionnaire statistics where available. 

\begin{figure}
    \centering
    \scalebox{0.95}{\includegraphics[]{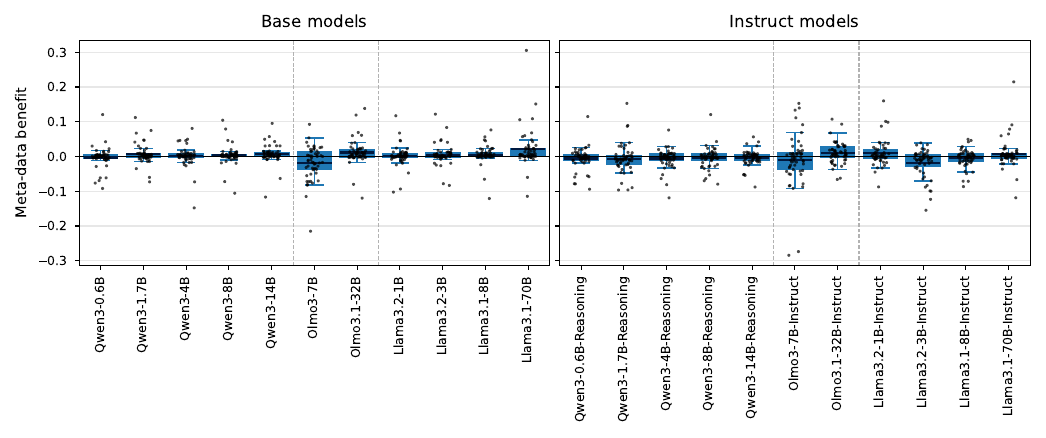}}
    \caption{Effect of persona-induction on individual-level behavioral prediction. Each subplot shows the effect of adding participant-specific meta-data to the prompt for a given model. Positive values indicate that persona-induction improves prediction, whereas negative values indicate that it worsens prediction.}
    \label{fig:fig5}
\end{figure}

In general, we found only a minor effect of persona-induction. The meta-data benefit ranged from $d = -0.02$ (Llama3.2-3B-Instruct) to $d = 0.02$ (Llama3.1-70B). This pattern held for both base and instruction-tuned models (see Fig.~\ref{fig:fig5}), and persisted when restricting the analysis to developmental experiments, where age-related differences in behavior are known to be present and informative. Taken together, these findings challenge the validity of persona-induction techniques, at least for modeling individual-level behavior. While such methods may change model outputs in ways that appear more human-like on the population level, they do not necessarily help to capture the behavior of individual people.

\subsection*{Discussion}

In this study, we have systematically evaluated the alignment between human behavior and LLMs. The results show that post-training consistently reduces behavioral alignment across model families, sizes, and post-training objectives. While base models still continue to improve across model generations, post-training misalignment actually increases in newer models, highlighting the pressing nature of this issue. These results were enabled by \textsc{Psych-201}, a dataset assembled through an open research collaboration that brought together contributions from across the behavioral sciences, yielding a resource of unprecedented scale. 

The observation that post-trained models exhibit reduced behavioral alignment with humans both connects to and extends several lines of earlier work. First, previous work has shown that early language models exhibited human-like cognitive biases, but that these patterns tend to disappear -- and were instead replaced with more rational behaviors -- in newer models that have undergone more extensive post-training \cite{hagendorff2023human, chen2023emergence, cheung2025large, shapira2026alignment}. Second, instruction‑tuned models have been found to produce less human‑like text than their base counterparts \cite{reinhart2025llms}. We extend this finding from text generation to behavioral responses, and demonstrate that it holds across a broad range of post‑training objectives. Lastly, looking beyond behavioral alignment, a similar observation has also been made in the psycholinguistic literature, where base models better predict human reading times than their instruction-tuned counterparts \cite{kuribayashi2024psychometric}. 

More broadly, our results can be viewed as a form of the alignment tax \cite{ouyang2022training} -- a phenomenon whereby post-training can degrade model capabilities acquired during pretraining. While recent work has proposed methods to mitigate such issues on common benchmarks \cite{lin2024mitigating}, our findings suggest that these mitigations do not extend to behavioral alignment with humans. It might thus be asked whether the observed misalignment reflects a fundamental limitation of post-training or merely shortcomings of current techniques. To probe this question, we evaluated Centaur, a model fine-tuned on a subset of tasks from \textsc{Psych-201} (i.e. those previously included in the \textsc{Psych-101} dataset). Centaur exhibited a consistent increase in behavioral alignment on held-out, novel tasks that were not part of its training set (average $d = 0.28$, SEM $= 0.10$), suggesting a substantial degree of generalization and demonstrating that post-training can, in fact, help.

These findings point to an important direction for future work: developing post-training methods that preserve the behavioral fidelity of base models while retaining the practical benefits of post-trained models. This includes not only capturing general patterns of human behavior, but also the stochasticity and individual-level variability that characterize human responses. The latter is especially critical when simulating behavior at the level of specific individuals or subpopulations. \textsc{Psych-201} provides rich participant-level metadata, which in principle allows us to build models that accomplish exactly this.

There is immense potential -- for both scientific and applied reasons -- for LLMs that closely mimic human behavior. If we had models that accurately simulate human learners, we could use them to design optimal educational curricula. If we had models that emulate human responses in mental health contexts, we could use them as training tools for psychiatrists. If we had models that accurately model public responses to policy changes, we could support more informed governance. To realize this potential, however, we need to move beyond the current paradigm in which post-training is exclusively optimized for normative correctness. \textsc{Psych-201} provides the foundation for evaluating behavioral alignment at scale, thereby directly paving the way towards this goal.


\subsubsection*{Funding}

This work was supported by the Helmholtz Association's Initiative and Networking Fund on the HAICORE@FZJ partition.  This work was funded by the German Research Foundation (DFG), Emmy-Noether grant ‘‘What’s in a name?’’ (project number 459717703), awarded to Fritz Günther. This work was funded by the German Research Foundation (DFG), research grant ‘‘A computational implementation of the Swinging Lexical Network model of language production’’ (project number 532390335), awarded to Fritz Günther. Evelina Leivada acknowledges funding by the Ministerio de Ciencia, Innovación y Universidades (Spain) under grant agreement CNS2023-144415. This work was made possible with the support of the NOMIS Foundation (Brian Christian,  Jian-Qiao Zhu, Evan M. Russek, and Thomas L. Griffiths).


\clearpage 

%
\bibliography{science_template} 
\bibliographystyle{sciencemag}

%
%
%
%
%
%



\newpage


\renewcommand{\thefigure}{S\arabic{figure}}
\renewcommand{\thetable}{S\arabic{table}}
\renewcommand{\theequation}{S\arabic{equation}}
\renewcommand{\thepage}{S\arabic{page}}
\setcounter{figure}{0}
\setcounter{table}{0}
\setcounter{equation}{0}
\setcounter{page}{1} 


\begin{center}
\section*{Supplementary Materials for\\ \scititle}

\end{center}

\subsubsection*{This PDF file includes:}
Materials and Methods\\
Figures S1 and S2\\
Study List\\

\newpage


\subsection*{Materials and Methods}

\subsubsection*{Data collection}

\textsc{Psych-201} was collected through an open research collaboration and crowdsourced effort. We issued a public call on social media inviting researchers to contribute datasets by submitting them to a GitHub repository. Each prompt was designed to include the entire trial-by-trial history of a complete session from a single participant. Human responses within these prompts were marked using the delimiters $<<$ and $>>$. For each dataset, 10\% of participants (or up to a maximum of 100) were reserved as a held-out test set. The submitted datasets underwent a lightweight review process to ensure that they did not contain obvious formatting or implementation bugs.

\subsubsection*{Large language models}

\begin{figure}[htb]
    \centering
    \includegraphics[]{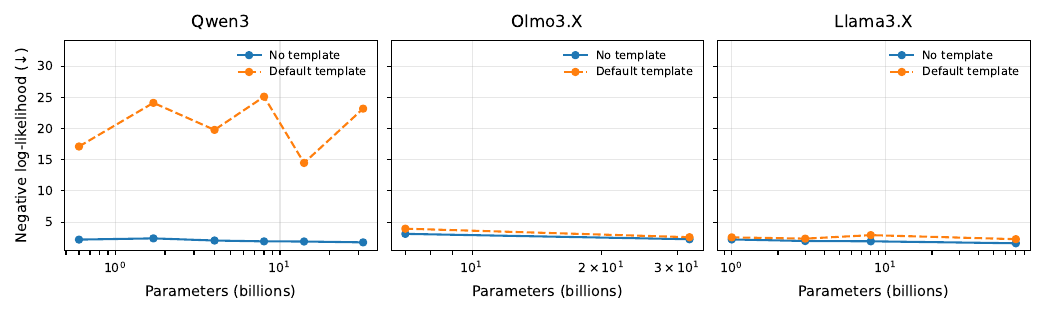}
    \caption{Behavioral alignment for instruction-tuned models prompted with and without chat template. Negative log-likelihood of human responses from \textsc{Psych-201} for models from the Qwen3, Olmo3.X, and Llama3.X families. Models prompted with their default chat template (orange dashed lines) consistently perform worse compared to those without a template (blue solid lines). The biggest gap is observed in the Qwen3 family.}
    \label{fig:s1}
\end{figure}

\begin{figure}[htb]
    \centering
    \includegraphics[]{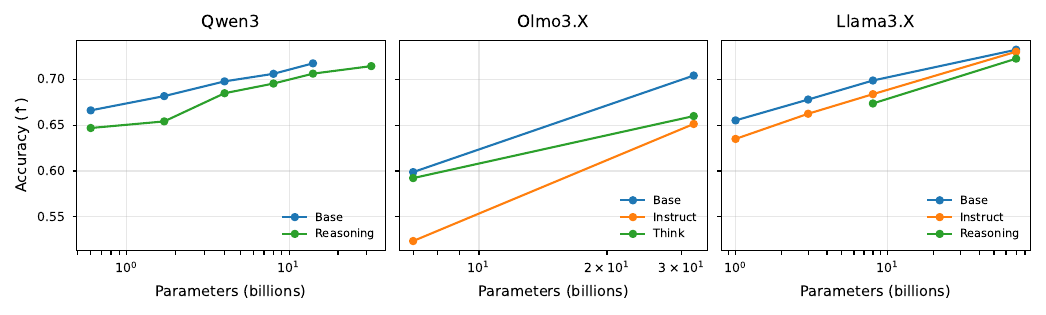}
    \caption{Behavioral alignment for different post-training objectives, measured by accuracy in predicting human responses on a discrete-choice subset of \textsc{Psych-201} for models from the Qwen3, Olmo3.X, and Llama3.X families. Post-trained models (shown in non-blue) exhibit consistently lower alignment with human responses across all objectives, families, and sizes than their base model counterpart (shown in blue). }
    \label{fig:s2}
\end{figure}

We conducted evaluations using three major families on \textsc{Psych-201}: Qwen3, a state-of-the-art open-source model family; Llama3.X, a model family with a broad ecosystem; and Olmo3.X, a fully open and reproducible model family. Models were run using the Hugging Face Transformers library in bfloat16 precision. The results reported in the main paper were obtained without applying a prompt template, as we found this setting to yield the best performance, including for models that provide a template. Further results obtained with the default model-specific templates are reported in Fig.~\ref{fig:s1}.

\subsubsection*{Evaluation metrics}

We used average negative log-likelihood over responses as our primary evaluation metric. Evaluation was restricted to the human response spans marked by the delimiters $<<$ and $>>$. For experiments in which a single response consisted of multiple tokens, we first summed the log-likelihoods across all tokens within that response and then averaged these values across responses.

We further quantified post-training misalignment by computing, for each experiment, Cohen’s d between the negative log-likelihoods of a base model and its post-trained counterpart, and then averaging the resulting effect sizes across experiments. We report all results on the held-out test set, whose size was sufficient to produce stable estimates.

\subsubsection*{Data availability}

Psych-201 is publicly available on the Huggingface platform at \url{https://huggingface.co/datasets/marcelbinz/Psych-201}. The test set is accessible under a CC-BY-ND-4.0 licence through a gated repository at \url{https://huggingface.co/datasets/marcelbinz/Psych-201-test}.

\subsubsection*{Code availability}

The code needed to reproduce our results is available at \url{https://github.com/marcelbinz/Psych-201}.

\newpage

\newpage

\subsection*{Study List (excluding \textsc{Psych-101})}

 {\footnotesize
  \begin{longtable}{@{}l l p{6cm}@{}}
  \label{tab:psych201}\\
  \toprule
  \textbf{Study} & \textbf{Domain} & \textbf{Task}\\
  \midrule
  \endfirsthead
  \toprule
  \textbf{Study} & \textbf{Domain} & \textbf{Task}\\
  \midrule
  \endhead
  \bottomrule
  \endlastfoot
  \texttt{aggarwal2023iag} \cite{aggarwal2022deceptive} & Reasoning & Inductive argument evaluation \\
  \texttt{agrawal2024stress} \cite{agrawal2023stress} & Intertemporal & Stress and delay discounting \\
  \texttt{akata2023repeatedgames} \cite{akata2025playing} & Econ. games & Repeated two-player games \\
  \texttt{alsobay2025publicGoodsGame} \cite{alsobay2026integrative} & Econ. games & Public goods game \\
  \texttt{anllo2024weird} \cite{anllo2024comparing} & Reinf. learning & Cross-cultural bandit task \\
  \texttt{anvari2024alien} \cite{anvari2024testing} & Reinf. learning & Bandit task (alien cover story) \\
  \texttt{anvari2024armed\_bandit} \cite{anvari2024testing} & Reinf. learning & Multi-armed bandit task \\
  \texttt{anvari2024observe\_bet} \cite{anvari2024testing} & Risky choice & Observe-or-bet task \\
  \texttt{anvari2024optional\_stopping} \cite{anvari2024testing} & Risky choice & Optional-stopping (secretary) task \\
  \texttt{anvari2024sampling\_paradigm} \cite{anvari2024testing} & Risky choice & Decisions from experience (sampling) \\
  \texttt{awad2018moral} \cite{awad2018moral} & Moral judgment & Moral Machine vehicle dilemmas \\
  \texttt{baar2022latent} \cite{van2022latent} & Econ. games & Social decisions; latent motives \\
  \texttt{barnby2022knowing} \cite{barnby2022knowing} & Econ. games & Social inference; intent attribution \\
  \texttt{bavard2018magnitude} \cite{bavard2018reference} & Reinf. learning & Instrumental learning; outcome magnitude \\
  \texttt{bavard2021range} \cite{bavard2021two} & Reinf. learning & Instrumental learning; range adaptation \\
  \texttt{bavard2023functional} \cite{bavard2023functional} & Reinf. learning & Value normalization in learning \\
  \texttt{bhatia2024likelihoodratings} \cite{bhatia2024exploring} & --- & Likelihood/probability ratings \\
  \texttt{binz2022heuristics} \cite{binz2022heuristics} & Reasoning & Heuristic multi-attribute choice \\
  \texttt{braendle2023empowerment} \cite{brandle2023empowerment} & Reinf. learning & Empowerment-driven exploration \\
  \texttt{breslav2022shuffle} \cite{breslav2022shuffle} & Reinf. learning & Reward learning task \\
  \texttt{burton2022optimism} \cite{burton2022optimism} & --- & Optimism bias in belief updating \\
  \texttt{busch2024\_navon} \cite{busch2024individual} & Cog. control & Navon global--local attention task \\
  \texttt{busch2024\_stroop} \cite{busch2024individual} & Cog. control & Stroop interference task \\
  \texttt{castro\_rodrigues2022twostep} \cite{castro2022explicit} & Reinf. learning & Two-step task \\
  \texttt{chambon2020feedback} \cite{chambon2020information} & Reinf. learning & Learning from free vs.\ forced choice \\
  \texttt{cheung2025omissionyesnobias} \cite{cheung2025large} & Moral judgment & Moral judgment of omissions \\
  \texttt{christian2025resolving} & Risky choice & Risky choice under uncertainty \\
  \texttt{ciranka\_vandenbos\_2024} \cite{ciranka2025internal} & Risky choice & Social influence on risk-taking \\
  \texttt{cohen2020causal} \cite{cohen2020rational} & Reasoning & Causal structure learning \\
  \texttt{decker2016twostep} \cite{decker2016creatures} & Reinf. learning & Two-step task across development \\
  \texttt{demircan2024evaluatingcategory} \cite{demircan2024evaluating} & Categorization & Category learning, naturalistic stimuli \\
  \texttt{demircan2024evaluatingreward} \cite{demircan2024evaluating} & Reinf. learning & Reward learning, naturalistic stimuli \\
  \texttt{dezfouli2019} \cite{dezfouli2019models} & Reinf. learning & Reward / reversal learning \\
  \texttt{dubois2022value} \cite{dubois2022value} & Reinf. learning & Value-free random exploration \\
  \texttt{evangelidis2023upscaling} \cite{evangelidis2023upscaling} & --- & Preference / choice task \\
  \texttt{fan2022trait} \cite{fan2023trait} & Reinf. learning & Reinforcement learning and traits \\
  \texttt{feher2020humans} \cite{feher2020humans} & Reinf. learning & Two-step task (model-based inference) \\
  \texttt{franke2024bayesian} \cite{franke2024bayesian} & Psycholinguistics & Bayesian pragmatic inference \\
  \texttt{frankedegen2016reasoning} \cite{franke2016reasoning} & Psycholinguistics & Pragmatic reasoning; implicature \\
  \texttt{frey2017dfe} \cite{frey2017risk} & Risky choice & Decisions from experience \\
  \texttt{frey2017lotteries} \cite{frey2017risk} & Risky choice & Choices between described lotteries \\
  \texttt{frey2017mpl} \cite{frey2017risk} & Risky choice & Risk elicitation (multiple price list) \\
  \texttt{gillan2016characterizing} \cite{gillan2016characterizing} & Reinf. learning & Model-based control and compulsivity \\
  \texttt{giron2023developmentalExploration} & Reinf. learning & Exploration across the lifespan \\
  \texttt{gross2023hindsightTransferLearning} \cite{gross2023knowledge} & --- & Transfer-learning task \\
  \texttt{guenther2020LDT} 2\cite{gunther2020trying} & Psycholinguistics & Lexical decision task \\
  \texttt{guenther2020TS} \cite{gunther2020semantic} & Psycholinguistics & Semantic similarity judgments \\
  \texttt{guenther2022relational} \cite{gunther2022patterns} & Psycholinguistics & Relational similarity judgments \\
  \texttt{guenther2023ViSpa} \cite{guenther2023vispa} & Psycholinguistics & Visual--semantic similarity judgments \\
  \texttt{guenther2023associations\_individual} & Psycholinguistics & Free word-association generation \\
  \texttt{guenther2023grammaticality} \cite{dentella2023systematic} & Psycholinguistics & Grammaticality judgments \\
  \texttt{guenther2024associations\_sentences\_texts} & Psycholinguistics & Word associations to texts \\
  \texttt{guenther2024comprehension} \cite{dentella2024testing} & Psycholinguistics & Text comprehension \\
  \texttt{guenther2024substitutions} \cite{pugacheva2024lexical} & Psycholinguistics & Word-substitution acceptability \\
  \texttt{gunadi2021deferral} \cite{gunadi2022impact} & Intertemporal & Choice deferral over time \\
  \texttt{haines2020intertemporal\_choice} \cite{haines2020anxiety} & Intertemporal & Delay-discounting choice \\
  \texttt{haridi2024memory} \cite{haridi2026context} & Memory & Episodic recognition memory \\
  \texttt{haridi2024memorySemanticContext} \cite{haridi2026context} & Memory & Recognition memory; semantic context \\
  \texttt{haridi2024memoryVisualContext} \cite{haridi2026context} & Memory & Recognition memory; visual context \\
  \texttt{hartley2024twoarmedbandit} \cite{nussenbaum2024sensitivity} & Reinf. learning & Two-armed bandit task \\
  \texttt{heffner2022\_economicgames} \cite{heffner2022probabilistic} & Econ. games & Social / economic games \\
  \texttt{hellmann\_unpublished\_brightness} & Psychophysics & Brightness discrimination \\
  \texttt{holton2024goalcommitment} \cite{holton2024goal} & Reinf. learning & Goal commitment and persistence \\
  \texttt{hu2023lmpragmatics} \cite{hu2023fine} & Psycholinguistics & Pragmatic language understanding \\
  \texttt{hunter2021increased} \cite{hunter2022increased} & Econ. games & Social decision-making \\
  \texttt{hussain2024risk} \cite{hussain2024novel} & --- & Risky choice task \\
  \texttt{jagadish2023zero} \cite{jagadish2023zero} & Reinf. learning & Zero-shot compositional generalization \\
  \texttt{jansen2021logic} \cite{jansen2021rational} & Reasoning & Logical reasoning problems \\
  \texttt{little2024functionestimation} \cite{little2025function} & --- & Function learning / estimation \\
  \texttt{marshall\_2022\_brightness} \cite{marshall2022magnitude} & Psychophysics & Brightness discrimination \\
  \texttt{moutoussis2018pavlovian} \cite{moutoussis2018change} & Reinf. learning & Pavlovian-to-instrumental transfer \\
  \texttt{nasioulas2024feedback} \cite{nasioulas2026feedback} & Risky choice & Risky choice with feedback \\
  \texttt{nussenbaum2020twostep} \cite{nussenbaum2020moving} & Reinf. learning & Two-step task across development \\
  \texttt{nussenbaum2023novelty} \cite{nussenbaum2023novelty} & Reinf. learning & Novelty-driven exploration \\
  \texttt{olschewski2024skewness} \cite{olschewski2024frequent} & Risky choice & Experience-based choice; skewness \\
  \texttt{olschewski2025optimal} \cite{olschewski2025optimal} & Risky choice & Information sampling in risky choice \\
  \texttt{palminteri2017confirmation} \cite{palminteri2017confirmation} & Reinf. learning & Instrumental learning; confirmation bias \\
  \texttt{phaneuf-hadd\_2025\_cogeff} \cite{phaneuf2025characterizing} & Cog. control & Cognitive effort allocation \\
  \texttt{pike2023catastrophizing} \cite{pike2023catastrophizing} & Risky choice & Sequential risk task \\
  \texttt{pirrone\_2018\_dots} \cite{pirrone2018single} & Psychophysics & Perceptual decision (dot discrimination) \\
  \texttt{pirrone\_unpublished\_food} & --- & Value-based choice (food) \\
  \texttt{pirrone\_unpublished\_lottery} & Risky choice & Choices between lotteries \\
  \texttt{potter2017twostep} \cite{potter2017cognitive} & Reinf. learning & Two-step task \\
  \texttt{rausch\_unpublished\_replication} & Reasoning & Reasoning task (replication) \\
  \texttt{rosenbaum2022valence} \cite{rosenbaum2022valence}  & Reinf. learning & Learning from good vs.\ bad outcomes \\
  \texttt{russek2024heuristics} \cite{russek2024heuristics} & Risky choice & Heuristics in sequential choice \\
  \texttt{rutledge2023happiness} \cite{rutledge2014computational} & Risky choice & Risky choice and momentary happiness \\
  \texttt{sandbrink2024metacontrol} \cite{sandbrink2026understanding} & Reinf. learning & Arbitration between control strategies \\
  \texttt{schiekiera2025metascience} \cite{schiekiera2025political} & Reasoning & Metascientific reasoning \\
  \texttt{shahar2019twosteptask} \cite{shahar2019improving} & Reinf. learning & Two-step task \\
  \texttt{singh2019phishing} \cite{singh2019training} & Reasoning & Phishing-email detection \\
  \texttt{singh2022representing} \cite{singh2022representing} & --- & Judgment / representation task \\
  \texttt{spektor2019contexteffects} \cite{spektor2019similarity} & Reinf. learning & Context effects in repeated choice \\
  \texttt{spektor2024lossaversion} \cite{spektor2024absolute} & Risky choice & Loss aversion, experience-based choice \\
  \texttt{sun2025rat} \cite{sun2025large} & Reasoning & Remote Associates Test (insight) \\
  \texttt{suthaharan2021paranoia} \cite{suthaharan2021paranoia} & Reinf. learning & Reversal learning and paranoia \\
  \texttt{tesslerfranke2018notunreasonable} \cite{tessler2018not} & Psycholinguistics & Interpretation of negated adjectives \\
  \texttt{thoma2025problearn} \cite{thoma2025emerging} & Reinf. learning & Probabilistic learning task \\
  \texttt{thoma2025riskychoice} \cite{thoma2025children} & Risky choice & Risky choice task \\
  \texttt{tsvilodub2023xorsome} \cite{tsvilodub2023role} & Psycholinguistics & Scalar implicature (or / some) \\
  \texttt{vandendriessche2022depression} \cite{vandendriessche2023contextual} & Reinf. learning & Reinforcement learning and depression \\
  \texttt{vantiel2021probabilisticpragmatics} \cite{van2021probabilistic} & Psycholinguistics & Probabilistic pragmatic inference \\
  \texttt{vantiel2022meaninguse} \cite{van2022meaning} & Psycholinguistics & Word meaning and use judgments \\
  \texttt{witte2024interventionStudy} \cite{witte2024exploring} & Reinf. learning & Exploration (intervention study) \\
  \texttt{witte2024safe\_exploration} \cite{witte2024exploring} & Reinf. learning & Safe exploration in bandits \\
  \texttt{witte\_thalmann2024exploration} \cite{witte2025model} & Reinf. learning & Exploration in bandit tasks \\
  \texttt{xu2021novelty} \cite{xu2021novelty} & Reinf. learning & Novelty-driven exploration \\
  \texttt{xu2023augmenting} \cite{xu2023augmenting} & Reasoning & Reasoning with augmentation \\
  \texttt{ying2023nipe} \cite{ying2023neuro} & Reasoning & Probabilistic inference task \\
  \texttt{zhu2024games} \cite{zhu2025capturing} & Econ. games & Economic games \\
  \texttt{zika2023} \cite{zika2023trait} & Reinf. learning & Learning under uncertainty \\
  \end{longtable}
  }
  
\end{document}